\documentclass[10pt,twocolumn,letterpaper]{article}

\usepackage[pagenumbers]{cvpr} % To force page numbers, e.g. for an arXiv version

\usepackage{graphicx}
\usepackage{amsmath}
\usepackage{amssymb}
\usepackage{booktabs}
\usepackage{multirow}

\usepackage[pagebackref,breaklinks,colorlinks]{hyperref}

\usepackage[capitalize]{cleveref}
\crefname{section}{Sec.}{Secs.}
\Crefname{section}{Section}{Sections}
\Crefname{table}{Table}{Tables}
\crefname{table}{Tab.}{Tabs.}

\def\cvprPaperID{*****} % *** Enter the CVPR Paper ID here
\def\confName{CVPR}
\def\confYear{2023}

\begin{document}

\title{Parallel Reasoning Network for Human-Object Interaction Detection}

\author{Huan Peng$^{1,2}$, Fenggang Liu$^2$, Yangguang Li$^2$, Bin Huang$^2$, Jing Shao$^2$, Nong Sang$^1$, Changxin Gao$^1$ \\
$^1$Huazhong University of Science and Technology \\
$^2$SenseTime Group \\
{\tt\small \{nsang,cgao\}@hust.edu.cn;} 
{\tt\small liyangguang@sensetime.com;} \\
{\tt\small \{penghuan,liufenggang,huangbin1,shaojing\}@senseauto.com}
}
\maketitle

%%%%%%%%% ABSTRACT
\begin{abstract}
Human-Object Interaction (HOI) detection aims to learn how human interacts with surrounding objects. Previous HOI detection frameworks simultaneously detect human, objects and their corresponding interactions by using a predictor. Using only one shared predictor cannot differentiate the attentive field of instance-level prediction and relation-level prediction. To solve this problem, we propose a new transformer-based method named Parallel Reasoning Network(PR-Net), which constructs two independent predictors for instance-level localization and relation-level understanding. The former predictor concentrates on instance-level localization by perceiving instances' extremity regions. The latter  broadens the scope of relation region to reach a better relation-level semantic understanding. Extensive experiments and analysis on HICO-DET benchmark exhibit that our PR-Net effectively alleviated this problem. Our PR-Net has achieved competitive results on HICO-DET and V-COCO benchmarks. 
\end{abstract}

%%%%%%%%% BODY TEXT
\section{Introduction}
\label{sec:intro}
The real world contains large amounts of complex human-centric activities, which are mainly composed of various human-object interactions (HOIs). In order for machines to better understand these complex activities, we need to detect all these HOIs accurately. To be specific, HOI detection can be defined as detecting the human-object pair and their corresponding interactions in an image. It can be divided into two sub-tasks, instance detection, and interaction understanding. Only if these two sub-tasks are completed can we build a good HOI detector.

\begin{figure}[t]
\begin{center}
\includegraphics[width=.48\textwidth]{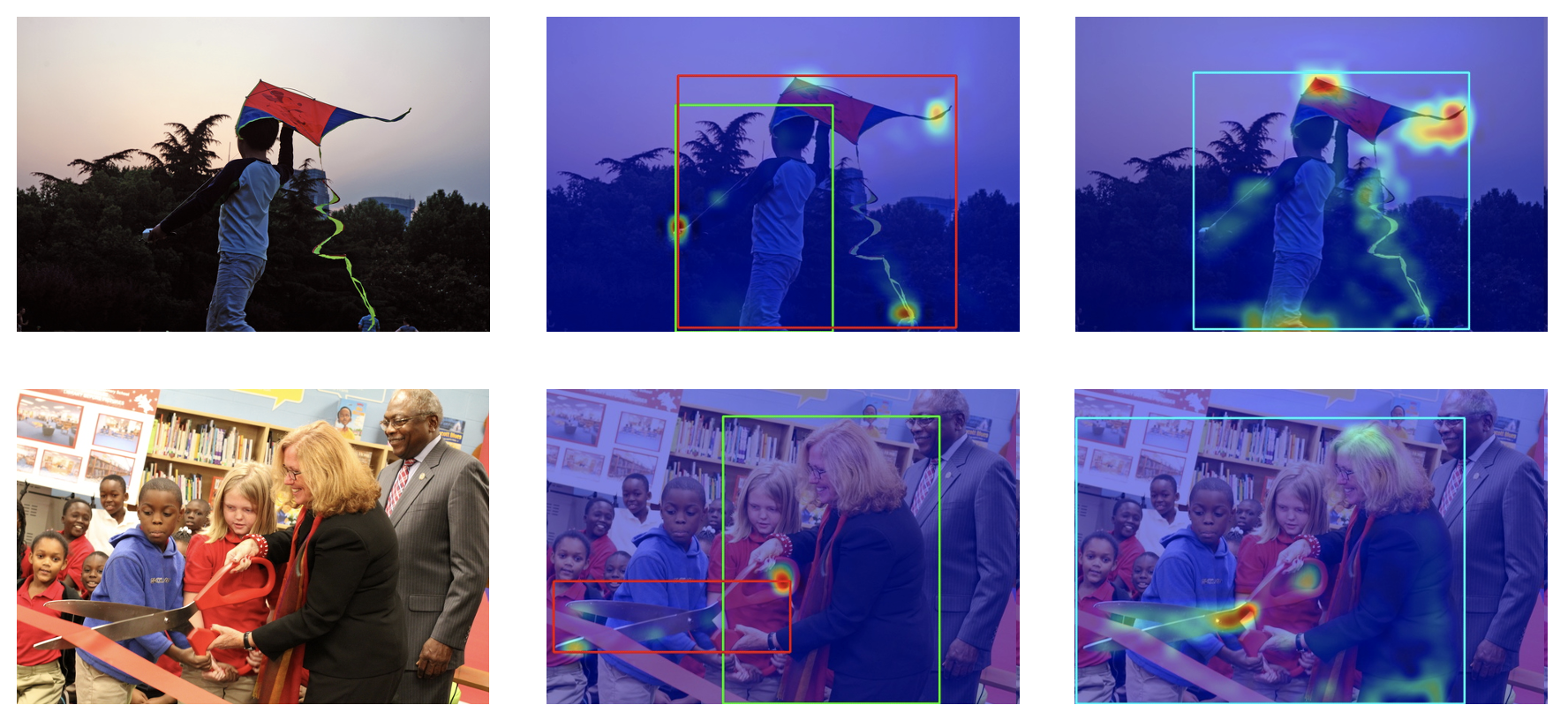}
\end{center}
   \caption{The attention fields for two different level predictors in our PR-Net. The first column shows these input images. The second column exhibits the attention fields of instance-level predictor, in which the model concentrates on the extremity region of human and object. The third column exhibits the attention fields of interaction-level predictor, in which the model spreads its scope of attention to the relation-level region.}
\label{fig1}
\end{figure}

Previously, different methods were taken to process these two sub-tasks.  Traditional methods like~\cite{hicodet,qi2018learning,no-frills,li2019transferable} first locates all instances and then extracts their corresponding features with an off-the-shelf object detector like~\cite{ren2017faster,he2017mask}. After that, instance matching and feature fusing approaches are used to construct human-object pairs which are more likely to have interactive relations. These pairs are then sent into the intention parsing network as inputs, and HOI is classified and outpus, so as to obtain the humain-object position and corresponding interactive relation category.  In summary, these traditional two-stage approaches suffer from the isolated training process of instance localization and interaction understanding, so they cannot localize interactive human-object pairs and understand those complex HOI instances.

To alleviate the above problems, multitask learning manners~\cite{liao2020ppdm,Wang2020IPNet,ggnet21,uniondet20,zou2021_hoitrans,chen_2021_asnet,tamura2021qpic,Kim_2021_CVPR} are proposed to complete these two sub-tasks simultaneously. Among these approaches, they~\cite{liao2020ppdm,Wang2020IPNet,ggnet21,chen_2021_asnet,Kim_2021_CVPR} process these two sub-tasks concurrently. Whereas they need an additional complex group composition procedure to match the predictions of these two sub-tasks, which 
reduces the computation efficiency. In addition, other one-stage methods~\cite{zou2021_hoitrans,tamura2021qpic} predict human-object pairs and corresponding interactions using one shared prediction head, without needing matching or gathering processes. However, they accomplish instance localization and interaction understanding in a mixed and tied manner. This naive mixed prediction manner can cause inconsistent focus in attentive fields between the instance-level and the relation-level prediction.
This inconsistent focus has caused limited interaction understanding for those hard-negative HOIs, which leads to dissatisfactory HOI detection performance.

To sum up, we propose a new transformer-based approach named Parallel Reasoning Network (PR-Net) to alleviate inconsistent focus of attentive fields for different level prediction. Specificly, two parallel predictos, instance-level predictor and relation-level predictor,are concluded in PR-Net. The former focuses on instance-level localization, and the latter keeps a watchful eye on relation-level semantic understanding. As can be seen from the two examples in the second columns of Figure~\ref{fig1}, PR-Net's attention to instances is focused on the endpoints of human skeleton and the particular edge regions of objects, indicating that the instance-level predictor can accurately locate the localization of human and objects by focusing on these critical extremity regions of instances. From the two examples in the third column of Figure~\ref{fig1}, it can be seen that PR-Net's attention to relational areas is focused on the interaction contact areas between human and objects and some contextual areas containing helpful understanding of the interaction, which indicates that the relational level predictor spreads its vision to relation areas to better understand the subtle relationships between human and objects. In addition, the instance-level queries of our instance-level predictor strictly correspond to the relation-level queries of our relationship-level predictors. So there is no need for any instance-level queries between them, which greatly reduces the computational cost~\cite{tamura2021qpic}.

Our contribution can be concluded in the following three aspects:
\begin{itemize}
    \item[$\bullet$] We propose PR-Net, which leverages a parallel reasoning architecture to effectively alleviate the problem of inconsistent focus in attention fields between instance-level and relation-level prediction. PR-Net achieves a better trade-off between two contradictory sub-tasks of HOI detection. The former needs more local information from the extremity region of instances, the latter is eager for more context information from the relation-level area.
    \item[$\bullet$] With  a decoupled prediction manner, PR-Net can detect various HOIs simultaneously without any matching or recomposition process to link the instance-level prediction and relation-level prediction.
    \item[$\bullet$] Equipped with additional techniques, including Consistency Loss for better training and Trident-NMS for better post-processing, PR-Net achieves competitive results on both HICO-DET and V-COCO benchmark datasets in HOI detection.
\end{itemize}

\section{Related Works}

\subsection{Two-stage Approaches in HOI Detection}
Most two-stage HOI detectors firstly detect all the human and object instances with a modern object detection framework such as Faster R-CNN, Mask R-CNN~\cite{ren2017faster,he2017mask}. After instance-level feature extraction and contextual information collection, these approaches pair the human and object instances for interaction recognition. In the process of interaction recognition, various contextual features are aggregated to acquire a better relation-level semantic representation. InteractNet~\cite{gkioxari2018detecting} introduces an additional branch for interaction prediction,  iCAN~\cite{gao2018ican} captures contextual information using attention mechanisms for interaction prediction. TIN~\cite{li2019transferable} further extends HOI detection models with a transferable knowledge learner.
In-GraphNet~\cite{in_GraphNet_ijcai2020} presents a novel graph-based interactive reasoning model to infer HOIs.
VSGNet~\cite{vsgnet} utilizes relative spatial reasoning and structual connections to analyze HOIs.
IDN~\cite{li2020hoi} represents the implicit interaction in the transformation function space to learn a better HOI semantic. 
Hou proposes fabricating object representations in feature space for few-shot learning~\cite{fcl} and learning to transfer object affordance for HOI detection~\cite{atl}. Zhang~\cite{scg} proposes to merge multi-modal features using a graphical model to generate a more discriminative feature.

\subsection{One-stage Approaches in HOI Detection}

One-stage approaches directly detect Human-Object Interactions without complicated coarse-to-fine bounding box regression~\cite{liao2020ppdm,Wang2020IPNet,ggnet21,uniondet20,zou2021_hoitrans,chen_2021_asnet,tamura2021qpic,Kim_2021_CVPR}. Among these approaches, ~\cite{liao2020ppdm,wang2020learning} introduced a keypoint-style interaction detection method which performs inference at each interaction key point. ~\cite{uniondet20} introduced a real-time method to predict the interactions for each human-object union box. Recently, transformer-based detection approach was proposed to handle HOI detection as a sparse set prediction problem~\cite{zou2021_hoitrans,chen_2021_asnet,tamura2021qpic}. Specifically, ~\cite{tamura2021qpic} designed a transformer encoder-decoder architecture to predict Human-Object Interactions in an end-to-end manner directly and introduced additional cost terms for interaction prediction. On the other hand, Kim \etal~\cite{hotr} and Chen \etal~\cite{asnet} propose an interaction decoder to be used alongside the DETR instance decoder. It is equally important for predicting interactions and matching related human-object pairs. These aforementioned one-stage approaches have enormously boosted the performance of Human-Object Interaction Detectors.

\begin{figure*}[htb]
\begin{center}
   \includegraphics[width=.7\linewidth]{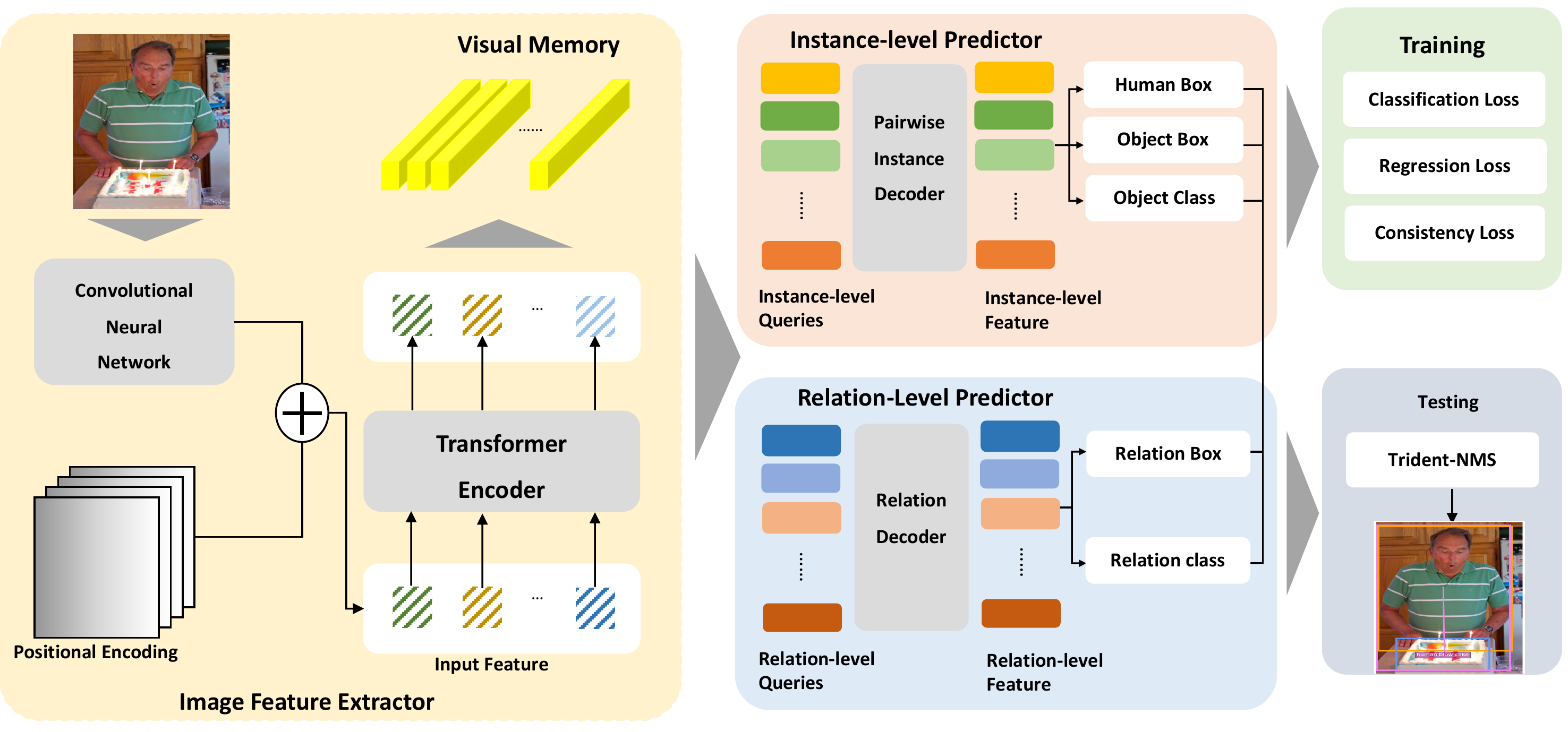}
\end{center}
\vspace{-1mm}
   \caption{The framework of our PR-Net. It is comprised of four components:\emph{Image Feature Extractor}, \emph{Pairwise Instance Predictor}, \emph{Relation-level Predictor}, \emph{Training and  Post-processing Techniques}.}
  \vspace{-2mm}
\label{fig:2}
\end{figure*}

\section{Proposed Method}
In this section, we present our Parallel Reasoning Network(PR-Net) for HOI detection, which is illustrated in the Figure~\ref{fig:2}. We can know that our PR-Net includes an Image Feature Extractor(CNN backbone and transformer encoder) and two parallel predictors (i.e., Instance-level Predictor and Relation-level Predictor). The two parallel predictors are designed to decode instance information(i.e. human-box, object-box, object-class) and relation information(i.e. relation-box, relation-class) respectively. Next, we introduce the proposed instance-level and relation-level loss functions to learn the location of instances and the interactions within each human-object pair. At last, we introduce the proposed Trident-NMS which is utilized to filter those duplicated HOI predictions effectively.

\subsection{Image Feature Extractor}
\label{img_extractor}
The overall Image Feature Extractor architecture consists of a standard CNN backbone $f_{c}$ and transformer encoder $f_{e}$. The conventional CNN backbone is used to process the input image $x \epsilon  \mathbb{R}^{3\times H \times W}$ to a global context feature map $z\epsilon  \mathbb{R}^{c \times H' \times W'}$, in which typically images are downsampled to $(H', W')$ spatial shape with a dimension of $c$. Then, the global context feature map is serialized as tokens, in which the spatial dimensions of the feature map are collapsed into one dimension, resulting in $H'\times W'$ tokens. Then, the tokens are linearly mapped to $\mathbf{T}=\{ t_i | t_i \epsilon  \mathbb{R}^{c'} \}_{i=1}^{N_q} $, where $N_q=H' \times W'$. Afterward, these tokens are shaped as a sequence to feed into the transformer encoder. 

For the transformer encoder, each encoder layer follows standard architecture of transformer, which consists of a multi-head self-attention module and a feed forward network (FFN). Additional position embedding $q_e \epsilon \mathbb{R}^{c' \times H' \times W' } $ is also added to the serialized token to supplement the positional information.  With the mechanism of self-attention, the encoder can map the former global context feature map from CNN to richer contextual information. Finally, the set of encoded image features $\{ d_i | d_i \epsilon  \mathbb{R}^{c'} \}_{i=1}^{N_q} $ can be formulated as visual memory $\mathbf{E}= f_{e}(\mathbf{T},q_e)$. The visual memory $\mathbf{E}$ contains richer contextual information.

\subsection{Instance-level Predictor}
\label{IP}
The Instance-level Predictor includes a standard transformer decoder $f_{ip}$ with just three layers. The decoder response for above visual memory $\mathbf{E}$, according to a set of learnable instance query vectors $\mathbf{Q_p}=\{ q_i | q_i \epsilon  \mathbb{R}^{c'} \}_{i=1}^{N_q} $ which is added with position embedding $p_l \epsilon \mathbb{R}^{c' \times H' \times W'} $. The instance-level queries vectors are trained to learn a more precise location of instances, which focuses more on those local information about location of instances.
The independent predictors are composed of three feed-forward networks (FFNs), including human-bounding-box FFN $\phi_{hb}$, object-bounding-box FFN $\phi_{ob}$, and object-class FFN $\phi_{oc}$, each of which response for decoding instance feature to human-box $\hat{b}^h$, object-box $\hat{b}^o$ and object-class $\hat{c}^o$  respectively. The formulation can be denoted as:

\begin{equation}
	\begin{aligned}
		\hat{b}^h &= \phi_{hb}(f_{ip}(\mathbf{Q_p}, p_l, \mathbf{E})), \\
		\hat{b}^o &= \phi_{ob}(f_{ip}(\mathbf{Q_p}, p_l,  \mathbf{E})), \\
		\hat{c}^o &= \phi_{oc}(f_{ip}(\mathbf{Q_p}, p_l, \mathbf{E})).
		\label{equ:ffn}
	\end{aligned}
\end{equation}

\subsection{Relation-level Predictor}
\label{RP}
We decouple the relation problems from HOI and use a Relation-level Predictor to reason relationships from larger-scale semantics. We propose a relation box to guide the predictor to percept the human-object relationship in the Relation-level Predictor.

The Relation-level Predictor consists of a standard transformer decoder $f_{rd}$ and two independent predictors(FFNs). Another relation-level queries $\mathbf{Q_r}$ and position embedding $p_r$ are randomly initialed and fed into the Relation-level Predictor. One of the predictors $\phi_{ub}$ predicts relation boxes $\hat{b}^u$, the other predictor $\phi_{ac}$ decodes the relation class information $\hat{c}^a$. The relation boxes $\hat{b}^u$ and the relation class information $\hat{c}^a$ can be formulated as Eq.~\ref{equ:ffn2}.

\begin{equation}
	\begin{aligned}
		\hat{b}^u &= \phi_{ub}(f_{dr}(\mathbf{Q_r}, p_r, \mathbf{E})), \\
		\hat{c}^a &= \phi_{ac}(f_{dr}(\mathbf{Q_r}, p_r, \mathbf{E})).
		\label{equ:ffn2}
	\end{aligned}
\end{equation}

\noindent Attributed to the relation boxes, the decoder of Interaction-level Predictor is guided to enlarge the receptive field (as shown in Figure~\ref{fig1}). The relation queries $\mathbf{Q_r}$ can pay attention to the entire area where human and object interact. Thus, the predictor $\phi_{ac}$  can predict a more accurate relation class.

In addiction, to match the relation class information $\hat{c}^a$ with the aforementioned human-box $\hat{b}^h$, object-box $\hat{b}^o$ and object-class $\hat{c}^o$ from the Instance-level Predictor, we ditch the complex matching method like HO pointer in HOTR. Instead, we just match the relation class information $\hat{c}^a$ and the instances information $\hat{b}^h$ \etc one by one in order. 
Specifically, for a pair of instances $\{\hat{b}^h_i,\hat{b}^o_,\hat{c}^o_i, i \epsilon N_q\}$, $\hat{c}^a_i$ is the corresponding relation class. In this way, the instance-level query vectors $\mathbf{Q_p}$ and the relation-level query vectors $\mathbf{Q_r}$ represent the same human-object interaction, but have the ability to focus on different receptive field. 
%Experiments show that  our simple matching method performs well.

\subsection{Loss Functions}
\label{loss_func}
The overall loss functions consist of the instance-level loss and relation-level loss, applied to Instance-level Predictor and Relation-level Predictor, respectively. The instance-level loss supervises the Instance-level Predictor to predict instance-level target, i.e., human-box, object-box, and object-class. The relation-level loss assists the Relation-level Predictor to predict relation-class and relation-box from the larger receptive field.

\subsubsection{The Instance-level loss function} $\mathcal{L}_{IL}$ supervises the instance information, including human-box $\hat{b}^h$, object-box $\hat{b}^o$ and object-class $\hat{c}^o$. The instance-level loss function consists of human-box regression $\mathcal{L}_{hr}$, object-box regression $\mathcal{L}_{or}$ and object-class classification $\mathcal{L}_{oc}$. $\mathcal{L}_{hr}$ and $\mathcal{L}_{or}$ are standard bounding-box regression loss, i.e. L1 loss, to locate the position of human and object.  $\mathcal{L}_{oc}$ is a classification loss to classify the categories of the object. The loss functions can be defined as Eq.~\ref{equ:bbox}.

\begin{equation}
	\begin{aligned}
		\mathcal{L}_{hr} &= \frac{1}{N} \sum_i^N {|| \hat{b}_i^h - b_i^h ||}, \\
		\mathcal{L}_{or} &= \frac{1}{N} \sum_i^N {|| \hat{b}_i^o - b_i^o ||}, \\
		\mathcal{L}_{oc} &= \frac{1}{N} \sum_i^N {\rm{CE}(\hat{c}^o_i, c^o_i)},
		\label{equ:bbox}
	\end{aligned}
\end{equation}

\noindent where CE is cross entropy loss, $c^o_i$ is the ground truth of object class. 

The instance-level loss function $\mathcal{L}_{IL}$  can be defined as:

\begin{equation}
	\begin{aligned}
		\mathcal{L}_{IL} = \mathcal{W}_{hr} *\mathcal{L}_{hr} + \mathcal{W}_{or} *\mathcal{L}_{or} + \mathcal{W}_{oc} *\mathcal{L}_{oc}.
		\label{equ:IL}
	\end{aligned}
\end{equation}

\subsubsection{The Relation-level loss function} $\mathcal{L}_{RL}$ supervises the relationship information, i.e., the relation class $\hat{c}^a$, primarily. In addition, auxiliary relation boxes are also supervised to pay attention to the entire area where the interaction happens. Thus, the Relation-level loss function consists of relation-box regression $\mathcal{L}_{ur}$, relation-box consistency loss $\mathcal{L}_{uc}$ and relation-class loss $\mathcal{L}_{ac}$. The $\mathcal{L}_{ac}$ is a classification loss to classify the categories of the interaction.
The relation-box regression loss function $\mathcal{L}_{ur}$ is a L1 loss to resemble the predicted relation boxes and its grounding truth. The grounding truth of relation boxes is the outer bounding box of human and object boxes. The relation-box regression loss function helps the Relation-level Predictor to be aware of the relation feature of human and object. 
The consistency loss $\mathcal{L}_{uc}$  are used to keep the consistency of $\hat{b}^h$, $\hat{b}^o$ and $\hat{b}^u$. Specifically, a pseudo relation box $\hat{b}^{ho}$ is generated by taking the outer bounding box of $\hat{b}^{h}$ and $\hat{b}^{o}$. Then, an L1 loss resemble $\hat{b}^u$ and $\hat{b}^{ho}$.
With the relation box, the relation-class loss can supervise better relation semantics.

\begin{equation}
	\begin{aligned}
		\mathcal{L}_{ur} &= \frac{1}{N} \sum_i^N {|| \hat{b}_i^u - b_i^u ||}, \\
		\mathcal{L}_{uc} &= \frac{1}{N} \sum_i^N {|| \hat{b}_i^u - \hat{b}_i^{ho} ||} , \\
		\mathcal{L}_{ac} &= \frac{1}{N} \sum_i^N {\rm{SigmoidCE}(\hat{c}^a_i, c^a_i)}. 
		\label{equ:bbox}
	\end{aligned}
\end{equation}

The relation-level loss function $\mathcal{L}_{RL}$  can be defined as:

\begin{equation}
	\begin{aligned}
		\mathcal{L}_{RL} = \mathcal{W}_{ur} * \mathcal{L}_{ur} + \mathcal{W}_{uc} *\mathcal{L}_{uc} + \mathcal{W}_{ac} *\mathcal{L}_{ac}.
		\label{equ:RL}
	\end{aligned}
\end{equation}

In all, the overall loss fucntion $\mathcal{L}$ can be denoted as:

\begin{equation}
	\begin{aligned}
		\mathcal{L} = \mathcal{L}_{IL} + \mathcal{L}_{RL}.
		\label{equ:all}
	\end{aligned}
\end{equation}

% \subsection{Testing and post-processing}
\subsection{Inference for HOI Detection}
\label{infer}
The inference process of our PR-Net can be divided into two parts: the calculation of the HOI predictions and the Trident-NMS post-processing technique. \\
\textbf{HOI Prediction} To acquire the final HOI detection results, we need to predict human bounding box, object bounding box, and object class using both instance-level predictions and relation class and relation box using relation-level prediction. Based on the above predictions, we can calculate the final HOI prediction score as below:
\begin{equation}
s^{hoi}_{i} = \{max_ks^o_{i}(k)\} * s^{rel}_{i}
\label{hoi_score}
\end{equation}
Where $max_ks^o_{i}(k)$ means the most probable class score of the $i$-th output object from instance-level predictor; $s^{rel}_{i}$ means the multi-class scores of the $i$-th output interaction from relation-level predictor. Note that each human-object pair can only have one object with certain class, but there maybe exist multiple human-object interactions within one pair. \\
\textbf{Trident-NMS} For each predicted HOI class in one image, we choose to filter its duplicated predictions according to the above calculated HOI prediction scores with our proposed Trident Non Maximal Suppression(Trident-NMS). In detail, if the $TriIoU(i, j)$ between the $i$-th and the $j$-th HOI prediction is higher than the threshold $Thres_{nms}$, we will filter the prediction which has a lower HOI score. And  the calculation of $TriIoU(i, j)$ is as below:

\begin{equation}
\begin{aligned}
     TriIoU(i, j) = & IoU(b^h_i, b^h_j) ^{W_h} \\
                    & \times IoU(b^o_i, b^o_j) ^{W_o} \\
                    & \times IoU(b^{rel}_i, b^{rel}_j) ^{W_{rel}} 
\end{aligned}
\label{tnms}
\end{equation}

Where $IoU(b^h_i, b^h_j)$, $IoU(b^o_i, b^o_j)$, $IoU(b^{rel}_i, b^{rel}_j)$ represent the Interaction over Union between the $i$-th and the $j$-th human boxes, object boxes and relation boxes; $W_h$, $W_o$, $W_{rel}$ represent the weights of Human IoU, Object IoU and Relation IoU.

\section{Experiment}

\subsection{Datasets and Evaluation Metrics}
\label{sec:dataset_metrics}
We evaluate our method on two large-scale benchmarks, including V-COCO \cite{gupta2015visual} and HICO-DET \cite{chao2018learning} datasets. V-COCO includes 10,346 images, which contains 16,199 human instances in total and provides 26 common verb categories. HICO-DET contains 47,776 images, where 80 object categories and 117 verb categories compose of 600 HOI categories. There are three different HOI category sets in HICO-DET, which are: (a) all 600 HOI categories (Full), (b) 138 HOI categories with less than 10 training instances (Rare), and (c) 462 HOI categories with 10 or more training instances (Non-Rare). 
Following the standard protocols, we use mean average precision $(mAP)$ in HICO-DET~\cite{hicodet} and role average precision $(AP_{role})$ in V-COCO~\cite{gupta2015visual} to report evaluation results.

\subsection{Implementation Details}
\label{sec:imple_detail}
We use ResNet-50 and ResNet-101\cite{resnet} as a backbone feature extractor. The transformer encoder consist of 6 transformer layers with multi-head attention of 8 heads. The number of transformer layers in Instance-level Predictor and Interaction-level Predictor is both set to be 3. The reduced dimension size of visual memory is set to $256$. The number of instance-level and relation-level queries is set to $100$ for both HICO-Det and V-COCO benchmark. 
The human, object and relation box FFNs both have $3$ linear layers with ReLU, while the object and relation category FFNs have one linear layer.
During training, we initialize the network with the parameters of DETR~\cite{carion2020endtoend} trained on the MS-COCO dataset.
We set the weight coefficients of bounding box regression, Generalized IoU, object class, relation class and consistency loss to $2.5$, $1$, $1$, $1$ and $0.5$, respectively, which follows QPIC~\cite{tamura2021qpic}. We optimize the network by AdamW~\cite{loshchilov2018decoupled} with the weight decay $10^{-4}$. We train the model for $150$ epochs with a learning rate of $10^{-5}$ for the backbone and $10^{-4}$ for the other parts decreased by $10$ times at the $100$th and the $130$th epoch respectively. All experiments are conducted on the $8$ Tesla A100 GPUs and CUDA11.2, with a batch size of $16$.

We select 100 detection results with the highest scores for validation and then adopt Trident-NMS to filter results further.

\subsection{Overall Performance}
\label{sec:perform}
We summarize the performance comparisons in this subsection.  

\textbf{Performance on HICO-DET.}
Table~\ref{tab:hicodet} shows the performance comparison on HICO-DET. Firstly, the detection results of our PR-Net are the best among all approaches under the Full and Non-Rare settings, demonstrating that our method is more competitive than the others in detecting the most common HOIs. It is noted that PR-Net is also preeminent in detecting rare HOIs (HOI categories with less than 10 training instances), because our parallel reasoning network can migrate the non-rare knowledge into a rare domain. Besides, our PR-Net obtains 32.86 mAP on HICO-DET (Default Full), which achieves a relative gain of 9.8\% compared with the baseline. These results quantitatively show the efficacy of our method.
% Performance on HICO-DET.

\begin{table}[htbp]
    \centering
    \caption{Results on HICO-DET~\cite{hicodet}. ``COCO'' is the COCO pre-trained detector, ``HICO-DET'' means that the detector is further fine-tuned on HICO-DET.} 
    \resizebox{0.45\textwidth}{!}{\setlength{\tabcolsep}{0.8mm}{
    \begin{tabular}{l c c | c c c}
        \hline
                                                 & & & \multicolumn{3}{c}{Default Full}     \\
        Method                                   & Detector & Backbone & Full & Rare & Non-Rare \\
        \hline
        \hline
        \textbf{CNN-based}  & & & & & \\
        VCL~\cite{vcl}                           & COCO     & ResNet-50     & 19.43 & 16.55 & 20.29 \\
        VSGNet~\cite{vsgnet}                     & COCO     & ResNet-152    & 19.80 & 16.05 & 20.91 \\
        DJ-RN~\cite{li2020detailed}                        & COCO     & ResNet-50     & 21.34 & 18.53 & 22.18 \\ 
        PPDM~\cite{liao2020ppdm}                         & HICO-DET & Hourglass-104 & 21.73 & 13.78 & 24.10 \\
        Bansal~$et~al.$~\cite{bansal2020}        & HICO-DET & ResNet-101 & 21.96 & 16.43 & 23.62 \\
        TIN~\cite{li2019transferable}$^{\rm{DRG}}$       & HICO-DET & ResNet-50      & 23.17 & 15.02 & 25.61 \\
        VCL~\cite{vcl}                           & HICO-DET & ResNet-50      & 23.63 & 17.21 & 25.55 \\
        GG-Net~\cite{ggnet21}	&HICO-DET	&Hourglass-104		&23.47	&16.48&	25.60\\
        IDN$^{\rm{DRG}}$~\cite{li2020hoi}                     & HICO-DET & ResNet-50      & 26.29 & 22.61 & 27.39  \\ 
        \hline
        \textbf{Transformer-based}  & & & & & \\
        HOI-Trans~\cite{zou2021_hoitrans}	& HICO-DET	&ResNet-50	&23.46& 16.91& 25.41       \\ 
        HOTR~\cite{Kim_2021_CVPR}	& HICO-DET	&ResNet-50		&25.10& 17.34 &27.42\\ 
        AS-Net~\cite{chen_2021_asnet}		& HICO-DET			&ResNet-50		&28.87	&24.25	&30.25   \\
        QPIC~\cite{tamura2021qpic}		& HICO-DET		&ResNet-50	&29.07	&21.85	&31.23   \\
        \textbf{PR-Net (Ours)}	& HICO-DET & \textbf{ResNet-50} & \textbf{31.17} & \textbf{25.66} & \textbf{32.82} \\
        \textbf{PR-Net (Ours)}~	& HICO-DET  & \textbf{ResNet-101} & \textbf{32.86} & \textbf{28.03} & \textbf{34.30} \\
        \hline
      \end{tabular}}}
        \label{tab:hicodet}
%  \vspace{-1cm}
\end{table}

\textbf{Performance on V-COCO.}
Comparison results on V-COCO in terms of \emph{$mAP_{role}$} are shown in Table~\ref{vcoco}. It can be seen that our proposed PR-Net has a {mAP(\%)} of 62.4, obtaining the best performance among all approaches. Although we do not adopt previous region-based feature learning (e.g., RPNN \cite{zhou2019relation}, Contextual Att \cite{wang2019deep}), or employ additional human pose (e.g., PMFNet \cite{wan2019pose}, TIN \cite{li2019transferable}), our method outperforms these approaches with sizable gains. Besides, our method achieves an absolute gain of 3.6 points, a relative improvement of 6.1\% compared with the baseline, validating its efficacy in the HOI detection task. 
% Performance on V-COCO.
\begin{table}[htbp]
\centering
\caption{Performance comparison on V-COCO dataset.}

\resizebox{0.45\textwidth}{!}{

\begin{tabular}{lccc}
\toprule
Method & Backbone Network & $\operatorname{AP}^{S1}_{role}$ & $\operatorname{AP}^{S2}_{role}$  \\ \hline
\textbf{CNN-based}   & & & \\
  VSGNet~\cite{vsgnet}  & ResNet-152  &51.8  &57.0\\ 
  PMFNet~\cite{wan2019pose}	  & ResNet-50-FPN  &52.0  &-\\ 
  PD-Net~\cite{zhong2020polysemy}   & ResNet-152-FPN  &52.6  &-\\
  CHGNet~\cite{wang2020contextual}	 & ResNet-50-FPN &52.7  &-\\
  FCMNet~\cite{fcm20}	 & ResNet-50  &53.1  &-\\
  ACP~\cite{kim2020detecting} & ResNet-152	&53.23  &-\\
  IDN~\cite{li2020hoi}	 & ResNet-50   &53.3  &60.3\\
  GG-Net~\cite{ggnet21}  & Hourglass-104 &54.7  &-\\
  DIRV~\cite{fang2020dirv}		 &  EfficientDet-d3  &56.1  &-\\
\midrule
\textbf{Transformer-based}  & & & \\
  HOI-Trans~\cite{zou2021_hoitrans}		 &  ResNet-101  &52.9  &-\\
  AS-Net~\cite{chen_2021_asnet}	 & ResNet-50  &53.9  &-\\
  HOTR~\cite{Kim_2021_CVPR} 	 & ResNet-50  &55.2  & 64.4\\
  QPIC~\cite{tamura2021qpic}  	 & ResNet-50 &58.8  &61.0 \\
\textbf{PR-Net (Ours)}~ & \textbf{ResNet-50} & \textbf{61.4} & \textbf{62.5} \\
\textbf{PR-Net (Ours)}~ & \textbf{ResNet-101} & \textbf{62.9} & \textbf{64.2}  \\
\bottomrule
\end{tabular}}

\label{vcoco}
\end{table}

\subsection{Ablation Analysis}
\label{sec:ablation}
To evaluate the contribution of different components in our PR-Net, we first conduct a comprehensive ablation analysis on the HICO-DET dataset. Next, we analyze the impact of the number of different-level predictors. At last, we analyze the effects of different post-processing manners.\\

\setlength{\tabcolsep}{4pt}
\begin{table}[htbp]
\begin{center}
  \caption{Ablation analysis of the proposed PR-Net with the backbone of ResNet-101 on HICO-DET test set. Parallel Predictor means we parallelly predict instance-level locations and relation-level semantics. Consistency Loss means we constrain the union box of the human-object pair and the relation box to be consistent. Trident-NMS means duplicate filtering through human, object, and relation bounding boxes.
  % Mean average precision (mAP) (\%) are reported.
   }
\small
\resizebox{0.45\textwidth}{!}{
\begin{tabular}{@{}ccccccc@{}}
\hline
\multirow{2}{*}{Parallel Predictor} & \multirow{2}{*}{Consistency Loss} & \multirow{2}{*}{Trident-NMS} &  \multicolumn{3}{c}{HICO-DET}\cr\cline{4-6}
% \cmidrule(lr){2-4} \cmidrule(lr){5-7}
&&&Full  &Rare &NonRare \cr
% Method & Full & Rare & NonRare \\
\hline\hline
- & - & - & 29.90 & 23.92 & 31.69 \\
\checkmark & - & - & 31.62 & 25.43 & 33.47 \\
\checkmark & \checkmark & - & 31.87 & 27.59 & 33.14 \\
\checkmark & \checkmark & \checkmark & 32.86 & 28.03 & 34.30 \\
\hline
\end{tabular}}

  \label{pr_3compo}
\end{center}
\end{table}
\setlength{\tabcolsep}{1.4pt}

\textbf{Contribution of different components.} Compared with our baseline~\cite{tamura2021qpic}, the performance improvements of our PR-Net are from three components: Parallel Predictor, Consistency Loss, and Trident-NMS. From Table~\ref{pr_3compo}, we can know the contribution of different components. Among these components, Parallel Predictor is our core approach. With that, we can observe a noticeable gain of mAP in HICO-DET by 1.72. It proves that the parallel reasoning structure can significantly improve instance localization and interaction understanding for an HOI detection model. Additionally, we design a consistency loss between the union box of the human-object pair and the relation box, which can contribute about 0.25 mAP gain in the HICO-DET test set. It shows that it is meaningful and helpful to constrain the union region of instance-level predictions and the relation region of relation-level predictions. At last, we design a more effective post-processing technique named Trident-NMS, which brings about 1.0 mAP gain in the HICO-DET test set. It reveals that the set-prediction method can also benefit from duplicate filtering technique and post-processing technique like NMS is essential for HOI detection.\\

\textbf{Impacts of different numbers of parallel predictors.} In our PR-Net, two parallel predictors are significant for HOI detection, and we detailedly analyze the impact of different numbers of parallel predictors. From Table~\ref{dec_num}, we can know that equipped with three layers of instance-level predictor and relation-level predictor, our PR-Net can acquire the best mAP performance in the HICO-DET test set. It reveals that our PR-Net can significantly outperform the baseline QPIC~\cite{tamura2021qpic} without additional computational cost. Interestingly, we can also observe that even with only one layer of parallel predictors, our PR-Net can also outperform the baseline equipped with a six-layer predictor.\\

\begin{table}[htbp]
\centering
\caption{ Ablation analysis of the number of instance-level predictor $N_{dec}$ and the number of relation-level predictor $N_{reldec}$.}
\resizebox{0.45\textwidth}{!}{
\begin{tabular}{lcccccc}
\toprule
approaches & Backbone    &   $N_{dec}$ &  $N_{reldec}$    & Full & Rare & Non-Rare \\ \hline
QPIC(Baseline)~\cite{tamura2021qpic}   & ResNet50     &   6   & -   & 29.07 & 
21.85 & 31.23 \\
PR-Net(Ours) & ResNet50     &   1   & 1   & 29.64 & 24.18 & 31.27 \\
PR-Net(Ours) & ResNet50     &   3   & 3   & \textbf{31.17} & \textbf{25.66} & 32.82 \\
PR-Net(Ours) & ResNet50     &   6   & 6   & 31.04 & 24.87 & \textbf{32.89} \\
\hline
QPIC(Baseline)~\cite{tamura2021qpic} & ResNet101    &   6   & -   & 29.90 & 23.92 & 31.69 \\
PR-Net(Ours) & ResNet101    &   1   & 1   & 30.26 & 23.27 & 32.34 \\
% PR-Net(Ours) & ResNet101    &   2   & 2   & 31.18 & 24.74 & 33.10 \\
PR-Net(Ours) & ResNet101    &   3   & 3   & \textbf{32.86} & \textbf{28.03} & \textbf{34.30} \\
PR-Net(Ours) & ResNet101    &   6   & 6   & 32.52 & 27.04 & 34.16 \\
\bottomrule
\end{tabular}}
\label{dec_num}
\vspace{-0.3cm}
\end{table}

\textbf{Effects of different implements of Trident-NMS.}
In Table~\ref{tnms_table}, we analyze the effects of different implements of Trident-NMS. We find that Product-based Trident-NMS performs better than Sum-based Trident-NMS. Additionally, we can also observe that when the weight of Human-IoU in $TriIoU$ increases, the HOI detection performance will be better. This reveals that human box duplication is more frequent than that of object box or relation box. In summary, with either the Product-based or Sum-based $TriIoU$ calculation, we should pay more attention to the non-maximal suppression of the human box.\\

\setlength{\tabcolsep}{4pt}
\begin{table}[h]
\begin{center}
\caption{Ablation analysis of the Trident-NMS module on HICO-DET test set.
Product means we calculate $TriIoU$ by multiplying these weighted Human-IoU, Object-IoU, and Relation-IoU.
Sum means we calculate $TriIoU$ by adding all these weighted Human-IoU, Object-IoU, and Relation-IoU.
$W_{h}$, $W_{o}$, $W_{rel}$ represent the weights of Human-IoU, Object-IoU and Relation-IoU respectively.
$Thres_{nms}$ means the threshold of non-maximum suppression.}
\small
\resizebox{0.45\textwidth}{!}{
\begin{tabular}{@{}ccccccccc@{}}
\hline
\multirow{2}{*}{Product} & 
\multirow{2}{*}{Sum} & 
\multirow{2}{*}{$W_{h}$}&
\multirow{2}{*}{$W_{o}$}& 
\multirow{2}{*}{$W_{rel}$}&
\multirow{2}{*}{$Thres_{nms}$}&
\multicolumn{3}{c}{HICO-DET}\cr\cline{7-9}
% \cmidrule(lr){2-4} \cmidrule(lr){5-7}
&&&&&&Full  &Rare &NonRare \cr

% Method & Full & Rare & NonRare \\
\hline\hline
- & -          & -   & -    & -   & -   & 31.87 & 27.59 & 33.14  \\
- & \checkmark & 0.33 & 0.33  & 0.33 & 0.5 & 30.61 & 27.00 & 31.69 \\
- & \checkmark & 0.33 & 0.33  & 0.33 & 0.7 & 32.53 & 27.88 & 33.91 \\
- & \checkmark & 0.4 & 0.4  & 0.2 & 0.7 & 32.63 & 27.96 & 34.02 \\
- & \checkmark & 0.5 & 0.4  & 0.1 & 0.7 & 32.66 & 27.91 & 34.00 \\
- & \checkmark & 0.6 & 0.3  & 0.1 & 0.7 & 32.56 & 27.70 & 34.01 \\
\hline
\checkmark & - & 1.0 & 1.0  & 1.0 & 0.5 & 32.77 & 27.98 & 34.20 \\
\checkmark & - & 1.0 & 1.0  & 0.5 & 0.5 & 32.81 & 28.02 & 34.25 \\
\checkmark & - & 0.5 & 0.5  & 0.5 & 0.5 & 32.61 & 27.65 & 34.08 \\
\checkmark & - & 0.5 & 1.0  & 0.5 & 0.5 & 32.61 & 27.67 & 34.09 \\
\checkmark & - & 1.0 & 0.5  &  0.5 & 0.5 & {\bf 32.86} & {\bf 28.03} & {\bf 34.30}  \\

\hline
\end{tabular}}
\label{tnms_table}
\end{center}
\end{table}
\setlength{\tabcolsep}{1.4pt}

\subsection{Visualization of features}
\label{sec:t_sne_sec}
\begin{figure}[h]
\begin{center}
\includegraphics[width=0.45\textwidth]{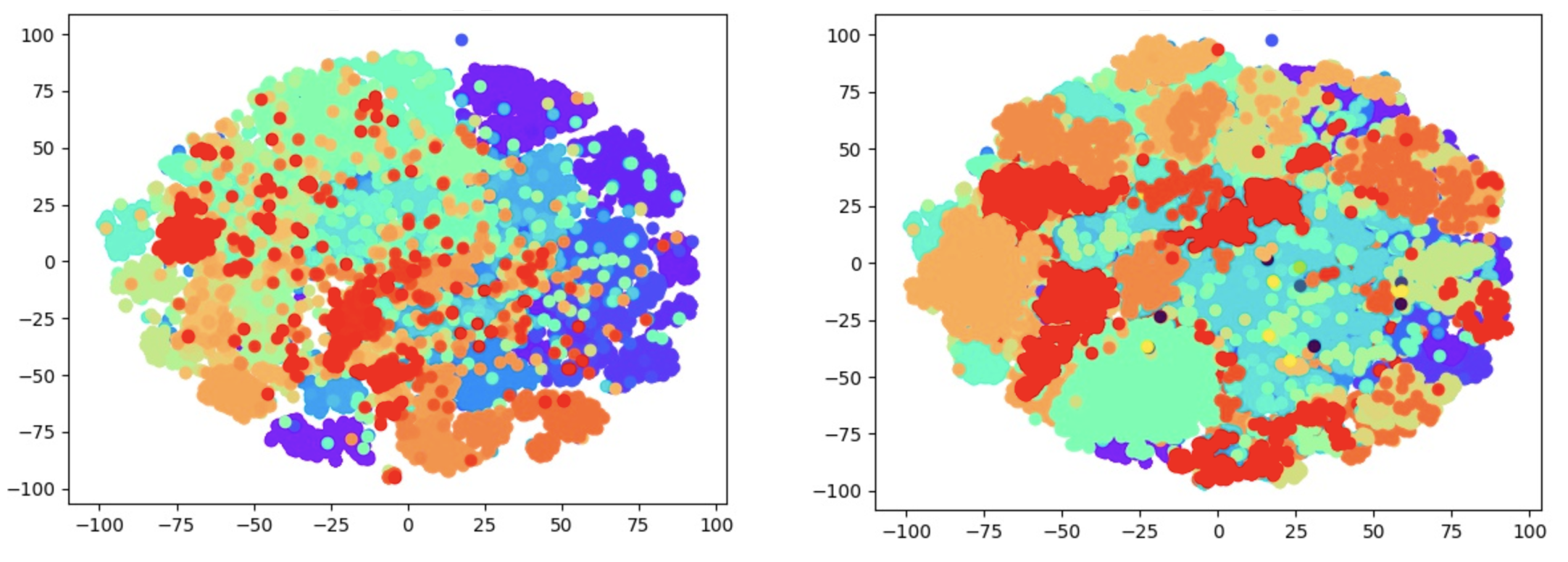}
\end{center}
\caption{Visualization of object features and relation features on HICO-DET dataset via t-SNE technique. Left is object features and right is relation features.}
\label{fig:t_sne}
\end{figure}

Using the t-SNE visualization technique~\cite{maaten2008visualizing}, we visualize 20000 samples of output feature. These object and interaction features are extracted from the last layer of Instance-level Predictor and Relation-level Predictor in our PR-Net, respectively. From the Figure~\ref{fig:t_sne}, we can observe that our PR-Net can obviously distinguish different class of objects and interactions. Interestingly, from this visualization of features, our PR-Net can even learn better the complex interaction representations then the object representations which benefits from our advantageous parallel reasoning architecture.

\subsection{Qualitative Examples}
\label{sec:quality}
\begin{figure}[h]
\begin{center}
\includegraphics[width=0.4\textwidth]{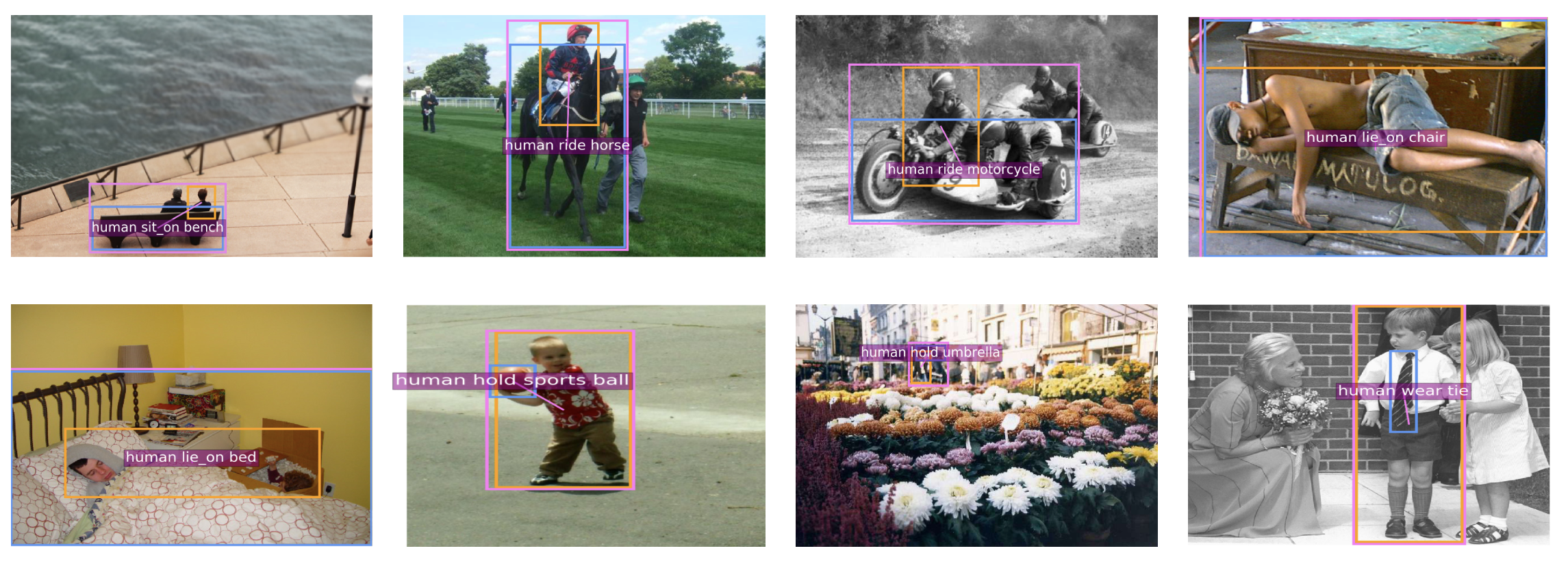}
\end{center}
\caption{Visualization of some HOI detection examples (Top 1 result) detected by the proposed Parallel Reasoning Network on the HICO-DET test set.}
\label{fig:visual_pics}
\end{figure}

From Figure~\ref{fig:visual_pics}, we can observe that our PR-Net can accurately detect both human box, object box, and relation box as well as their corresponding interactions. From the first row and second column of Figure~\ref{fig:visual_pics}, we can know that our PR-Net can precisely distinguish which man is riding the horse in the image. From the second row and third column of Figure~\ref{fig:visual_pics}, our PR-Net can precisely detect those subtle and indiscernible HOIs. In summary, our PR-Net can correctly detect those complex and hard HOIs. 

%%% noise
\section{Conclusion}

In this paper, we propose a new Human-Object Interaction Detector named Parallel Reasoning Network(PR-Net), which consists of an instance-level predictor and a relation-level predictor, to alleviate the problem of inconsistent focus in attentive fields between instance-level and interaction-level predictions. In addition, our PR-Net achieves a better trade-off between instance localization and interaction understanding. Furthermore, equipped with Consistency Loss and Trident-NMS, our PR-Net has achieved competitive results on two main HOI benchmarks, validating its efficacy in detecting Human-Object Interactions.

\clearpage

%%%%%%%%% REFERENCES
{\small
\bibliographystyle{ieee_fullname}
\bibliography{egbib}
}

\end{document}